# Квантитативна параметризація текстів Івана Франка: спроба проекту

*Соломія Бук*

Кандидат філол. наук, доцент кафедри загального мовознавства
Львівського національного університету імені Івана Франка
вул. Університетська, 1, Львів 79000,
тел. 239-47-56, факс. 297-16-68, e-mail: solomija@gmail.com

У статті заманіфестовано проект квантитативної параметризації усіх текстів І. Франка, що можливо реалізувати, створивши частотний словник усіх творів письменника і лише із застосуванням сучаних комп'ютерних розробок. Вказано сфери застосування, етапи, методику, принципи і специфіку укладання частотного словника мови другої половини XIX — поч. XX ст., якою писав І. Франко. Описано співвідношення частотного словника І. Франка із словником мови письменника та корпусом текстів.

Ключові слова: частотний словник (ЧС), словник мови письменника, ідіостиль І. Франка, лінгвостатистичний аналіз тексту, корпус текстів.

Якщо зіставити кількість літературознавчих та мовознавчих досліджень творчості Івана Франка, то шалька терезів схилиться у сторону перших. Лінгвістичні ж напрямки аналізу доробку письменника торкались багатьох питань: ролі Івана Франка в творенні літературної мови[1], внеску в ономастику[2], розуміння взаємозв'язку мови і духовності[3], деяких проблем загального мовознавства у його науковій спадщині[4] та термінології[5]. Чи не найширшого висвітлення отримала стилістика творів[6], зокрема наголосу в поезії[7], вживання діалектизмів[8], арготизмів[9], синонімів[10], дослідження синтаксису творів[11]. Зазначимо, що творчість Івана Франка привертає увагу не лише українських науковців. Наприклад, результати дослідження довжини слів у листах і поетичних творах Івана Франка наведено в німецькомовних працях[12].

---

[1] *Франко З.Т.* Мова творів Івана Франка. // Курс історії української літературної мови. – Т. 1. – К.: Вид-во АН УРСР, 1958. – С. 476-519; *Гузар О.* Іван Франко і становлення єдиного українського правопису // Іван Франко — письменник, мислитель, громадянин. Матеріали міжнародної наукової конференції (Львів 25-27 вересня 1996) / Ред. *М. Ільницький, Б. Якимович.*— Львів: Світ, 1998.— С. 721–722.

[2] *Vlasenko-Bojchun A.* Franko's contribution to onomastics // Іван Франко — мислитель і митець: Зб. доп. для відзначення 125-річчя народин і 85-річчя смерті Івана Франка.— Нью-Йорк; Париж; Сідней; Торонто, 1981.— С. 120–127.

Проте, незважаючи на згадане широке коло опрацьованих питань у творчості Генія, залишаються майже невирішені. Серед них, зокрема, — створення словника мови І. Франка, анотованого корпусу його текстів, а також квантитативне дослідження творів письменника. У лінгвістиці зламу XX–XXI століть особливо актуальними стали міждисциплінарні дослідження мови, до яких належить й застосування статистичних методів до мовних об'єктів.

---

[3] *Сербенська О.А.* Основи мовотворчості журналіста в інтерпретації Івана Франка: Текст лекцій. – Львів: Ред.-вид. відділ Львів. ун-ту, 1992. – 112 с.

[4] *Ковалик І. І.* Наукова лінгвістична проблематика в працях Івана Франка // Іван Франко. Статті і матеріали.— Зб. 12.— К., 1965.— С. 113 – 121; *Мацюк Г. П.* І. Франко про вирішення правописних питань у Галичині // Іван Франко і національне відродження.— Львів: ЛДУ ім. І Франка; Ін-т франкознавства, 1991.— С. 187–188; *Галенко І.* Проблеми загального мовознавства в науковій спадщині Івана Франка // Іван Франко — письменник, мислитель, громадянин.— С. 691–700; *Медвідь А.* Іван Франко: поняття рідної мови // Іван Франко — письменник, мислитель, громадянин.— С. 682–686.

[5] *Кочан І.* Термінологічні проблеми у працях Івана Франка // Іван Франко — письменник, мислитель, громадянин.— С. 701–703.

[6] *Полюга Л. М.* Слово у поетичному тексті Івана Франка.— К.: Наук. думка, 1977.— 167 с.; *Бацевич Ф.* Імпліцитні текстові засоби в повістях Івана Франка // Іван Франко — письменник, мислитель, громадянин.— С. 662–666; *Ціхоцький І. Л.* Мовна характеристика персонажів у прозі Івана Франка: Автореф. дис. … канд. філол. наук. — Львів, 2004. — 20 с.

[7] *Ovcharenko M.* Stress in Ivan Franko's poetry // Annals of the Ukrainian Academy of Arts and Science in U. S.— 1960.— Vol. 8.— № 1–2 (25–26).— P. 121–140.

[8] *Закревська М.* Внесок Івана Франка у розвиток науки про українські діалекти // Іван Франко — письменник, мислитель, громадянин.— С. 652–656; *Лев В.* Західно-українські елементи мови в ранній творчості Івана Франка // Іван Франко — мислитель і митець: Збірка доповідей для відзначення 125-річчя народин і 85-річчя смерті Івана Франка / Редакція *Є. Федоренка*.— Нью-Йорк; Париж; Сідней; Торонто, 1981.— XII; *Рабій-Карпинська С.* Говори Дрогобиччини з узглядненням говірки села Нагуєвичі, Івано-Франківська область // Іван Франко — мислитель і митець.— С. 128–138.

[9] *Горбач О.* Вулично-тюремні арготизми у Франковій прозі // Записки НТШ.— 1963.— Т. 177 (117).— С. 197–206.

[10] *Ощипко І.* До вивчення прикметникової та прислівникової синоніміки в художніх творах І. Я. Франка // Питання українського мовознавства.— Кн. 4.— Львів, 1960.— С. 97–103.

[11] *Петличний И. З.* Синтаксис языка призведений Ивана Франко. На материале художественной прозы: Автореферат дисс. … докт. филолог. наук.— Львов, 1965.— 32 с.

[12] *Best K.-H., Zinenko S.* Wortkomplexität im Ukrainischen und ihre linguistische Bedeutung // Zeitschrift für Slavische Philologie.— 1998.— B. 58.— S. 107–123; *Best K.-H., Zinenko S.* Wortlängen in Gedichten des ukrainischen Autors Ivan Franko // Pange lingua. Zborník na počesť Viktora Krupu / *J. Genzor, S. Ondrejovič*.— Bratislava: Veda, 1999.— S. 201–213.

Із цього погляду особливої уваги заслуговує задум створення "Словника мови Івана Франка" науковцями Львівського університету під керівництвом проф. І. Ковалика. Зважаючи на величезний обсяг творчості І. Франка в різних родах і жанрах, було вирішено видати спочатку словник поезії. Першою фазою створення лексикону справедливо визначили укомплектування реєстру слів. Проф. І. Ковалик розробив принципи та наукові основи укладання словника[13], базуючись на яких було видано пробний зошит мови художніх творів І. Франка[14]. Карткування віршованої спадщини відбувалося вручну, то ж можна тільки уявити, скільки роботи виконано для виявлення лише списку слів кількістю понад 35 000, що побачив світ у книзі "Лексика поетичних творів..."[15]. "Мета цього реєстру на основі підрахунків, наявних у картотеці слів, подати всю українську лексику поетичних творів Івана Франка та вказати на частотність використання кожного слова"[16]. У словнику свідомо збережено діалектні, фонетичні, морфологічні і граматичні написання, матеріал подано єдиним списком за алфавітом. На основі словопокажчика поезій І. Франка Л. Полюга здійснив короткий статистичний аналіз лексики його поезій[17], а І. Ощипко порівняла дані зі словниками мови Т. Шевченка та Г. Квітки-Основ'яненка[18].

Вказане видання залишається досі, на жаль, єдиним лексикографічним опрацюванням ідіостилю І. Франка, що дає підстави вважати проблему створення словника мови цього письменника відкритою. Коротко оглянемо словники, об'єктом опису яких є текст письменника.

---

[13] *Ковалик І. І.* Принципи укладання Словника мови творів Івана Франка // Українське літературознавство. Іван Франко. Статті і матеріали.— Львів, 1968.— Вип. 5.— С. 174–183.; *Ковалик І. І.* Наукові філологічні основи укладання і побудови Словника мови художніх творів Івана Франка // Українське літературознавство. Іван Франко. Статті і матеріали.— Львів, 1972.— Вип. 17.— С. 3–10.

[14] *Ковалик І. І.* Словник мови художніх творів Івана Франка. Пробний зошит // Українське літературознавство. Іван Франко. Статті і матеріали.— Львів, 1976.— Вип. 26.— С. 63–99.

[15] *Ковалик І. І., Ощипко І. Й., Л. М. Полюга* (уклад.) Лексика поетичних творів Івана Франка. Методичні вказівки з розвитку лексики.— Львів: ЛНУ, 1990.— 264 с.

[16] Передмова // Там само.— С. 4.

[17] *Полюга Л.* Статистичний аналіз лексики поетичних творів І. Франка // Іван Франко і національне відродження.— Львів: ЛДУ ім. І Франка; Ін-т франкознавства, 1991.— С. 164–166.

[18] *Ощипко І.Й.* Про укладання словника мови поетичних творів Івана Франка // Іван Франко і світова культура. Матеріали Міжнар. симпозіуму ЮНЕСКО (Львів, 11-15 вересня 1986).— Кн. 1.— К.: Наук. думка, 1990.— С. 82.

## Огляд українських словників, об'єктом опису яких є текст письменника

У сучасній лексикографії словники, об'єктом опису яких є текст письменника, відповідно до мети і можливостей укладачів представлені різними типами: словник мови письменника, словопокажчик, конкорданс, частотний словник та порівняно новий тип — лексична будова ідіолекту[19].

Найповніший опис лексико-фразеологічного складу творів пропонує словник мови письменника[20]. Цей модифікований тип словника поєднує тлумачну та письменницьку лексикографічні праці. "Цей словник має значення як для вивчення творчості самого письменника, так і для дослідження відповідного етапу історії національної літературної мови та суспільної атмосфери того періоду"[21]. Суттєвою рисою словника мови письменника є те, що він подає тільки ті значення, в яких слово реалізовано у творах письменника. За таким принципом описано лексикони Т. Шевченка[22] та Г. Квітки-Основ'яненка[23].

Словопокажчик вказує на том, сторінку і в деяких випадках на рядок, де функціонує слово у конкретній словоформі, а також на частоту уживання кожної одиниці. В Україні вийшли друком словопокажчики до творів

---

Т. Шевченка[24], І. Котляревського[25], Лесі Українки[26], А. Тесленка[27], В. Стефаника[28], Ю. Федьковича[29].

Глосарій — зібрання глос, тобто незрозумілих, з погляду укладачів, слів та пояснень чи коментарів до них. До цього лексикографічного типу можна зарахувати "Словник поетичної мови Василя Стуса"[30], де подано тлумачення складної для розуміння лексики з творів письменника, а саме авторських неологізмів, архаїзмів та історизмів, діалектизмів та інших рідковживаних слів.

Конкорданс подає усі або вибіркові контексти вживання певного слова чи словоформи, наприклад, конкорданс поетичних творів Тараса Шевченка "точно відмічає, де в "Кобзарі", і в яких текстуальних обставинах появляється кожна вжита Шевченком словоформа, як українська, так і російська"[31]. У ньому зафіксовано 18 401 лексичну одиницю і подано 83 731 випадки їхнього вживання. Опрацювання матеріалу здійснено з використанням комп'ютерних програм, на відміну від описаних вище праць.

Частотний словник (далі — ЧС) — тип словника, де наведено кількість вживань, тобто *частоту* певної одиниці мови в обстежених текстах. Звичайно він складається з декількох списків: списків слів, зведених до початкової форми, розташованих за спадом частот і за алфавітом, а також з таких самих списків для словоформ. Мовознавці зазначають, що "... ЧС є важливим інструментом дослідження закономірностей функціонування лексичної системи

---

мови в різних текстах..."[32]. В українській лексикографії існують ЧС шести функціональних стилів[33], проте аналога ЧС письменника чи його окремого твору як, наприклад, ЧС "Войны и мира" Л. Толстого[34], ще не створено.

Усі перелічені вище типи словників тісно пов'язані між собою, так, скажімо, укладаючи словник мови письменника, для розрізнення значень слів та їхньої ілюстрації послуговуються контекстними даними конкордансу. Словникову статтю глосарія будують за таким самим зразком, що й у словнику мови письменника, тільки реєстр слів добирають за іншими критеріями. Той факт, що словопокажчик і словник мови подають, як правило, кількість уживань кожного слова, зближає їх із частотним словником. Таким чином, можна констатувати перше застосування статистичних методів до виявлення лексичних особливостей творів української літератури зі середини ХХ ст. у словниках мови письменників та у словопокажчиках.

Проте є декілька суттєвих рис, на перший погляд, формальних, але насправді принципових, що не дозволяють прирівнювати частотний словник із вказаними працями — це послідовність розміщення словникових статей. Для того, що би словник можна було вважати частотним, він обов'язково повинен містити ранговий список одиниць, тобто список, у якому першим стоїть слово з найбільшою частотою, на другому — друге за частотністю і т. д. Номер слова за порядком називається рангом. Така форма подання інформації дає можливість визначити обсяг словника слів і словоформ, величину покриття тексту, багатство словника, індексів винятковості й концентрації та інше. Словник мови письменника і словопокажчик подають реєстр слів за алфавітом, що не дає можливості не те що обчислити важливих статистичних характеристик тексту, а навіть визначити найчастотніших одиниць, заради чого, зокрема, і укладаються ці лексикографічні праці. Ця проблема поглиблюється ще й тим фактом, що вказані словники існують лише в паперовій формі, і виявити найчастотнішу лексику можливо тільки після громіздкої ручної роботи. Для цього доведеться опрацювати все видання і вручну шукати й виписувати слова за спадом частот.

---

[32] *Перебийніс В. С., Муравицька М. П., Дарчук М. П.* Частотні словники та їх використання.— К.: Наукова думка, 1985.— С. 78.

[33] Їх огляд див.: *Бук С. Н.* Лексична основа української мови: виділення та системно-структурна організація.— Рукопис. Дис... канд. філол. наук: 10.02.01 / Львівський національний університет імені Івана Франка.— Львів, 2004.— С. 31.

[34] *Великодворская З. Н. и др.* (ред.) Частотный словарь романа Л. Н. Толстого "Война и мир" / М-во просвещения РСФСР. —Тула: Б. и., 1978. — 380 с.

Для чого ж потрібні і як можна використовувати частотні словники певного твору чи письменника? Спробуємо коротко окреслити теоретичні та практичні сфери застосування ЧС.

**Сфери застосування частотного словника творів письменника**

1. Насамперед, природна мова, зокрема текст, має власні квантитативні закономірності, тому лінгвістичні дослідження із врахуванням лише якісної її оцінки буде нерізностороннім і неповним. "Шлях дисципліни вглиб рано чи пізно наштовхується неминуче на обмеженість якісних методів, на безпорадність неточного способу вираження, на відсутність гіпотез, а також на відсутність теорії," — зауважив знаний німецький лінгвіст Ґ. Альтманн[35].

Те, що словесне наповнення будь-якого достатньо довгого тексту має власну статистичну структуру, доведено ще на початку XX ст. Вона виявляється у тому, що розподіл частоти одиниць мови в тексті має певну регулярність і може бути описаний за допомогою певних моделей і теоретичних формул, наприклад, мова і мовлення надають перевагу невеликій кількості одиниць, які часто використовуються і становлять ядро будь-якої мовної чи мовленнєвої підсистеми, тоді як переважна кількість одиниць є низькочастотними (закон переваги)[36]. Екстраполюючи це правило на словник письменника, можна стверджувати, що в кожного автора існує строге співвідношення більш і менш частотних лексем. Різниця між статистичною структурою текстів є критерієм унаочнення відмінностей між ними, тобто відмінностей між стилями порівнюваних письменників[37]. "Статистичний аналіз сучасних українських текстів, стильових різновидів української мови засвідчив, що вони відрізняються одиницями всіх рівнів мови, але стилерозрізнювальна потужність цих одиниць різна: найнижчою вона є на фонемному рівні, найвищою — на лексичному та синтаксичному"[38]. Власне статистичну структуру лексичного рівня певного твору чи мови письменника і виявляє ЧС.

---
[35] *Альтман Ґ.* Мода та істина в лінгвістиці // Проблеми квантитативної лінгвістики.— Чернівці: Рута, 2005.— С. 3–11.— С. 6.
[36] *Перебийніс В. С.* Статистичні методи для лінгвістів: Навчальний посібник.— Вінниця: "Нова книга", 2002.— С. 7.

2. Зазначене співвідношення більш і менш частотних лексем у певних письменників читач інтуїтивно сприймає як різноманітний чи одноманітний словниковий запас творів. У науковій літературі (як літературознавчій, так і мовознавчій) часто у дослідженнях оперують такими фразами як "лексика цього письменника багатша за лексику іншого", "цей письменник використовує більше епітетів, ніж той" і т. д. Ці твердження залишаються голослівними (або відносними, суб'єктивними, недоведеними) до того часу, поки не здійснено конкретних досліджень, зокрема порівнянь співвідношення частин мови у творах певного письменника з аналогічним співвідношенням в творах іншого, а також таких статистичних характеристик лексикону як багатство словника чи індекс різноманітності (відношення обсягу словника лексем до обсягу тексту), індекс винятковості (відношення кількості слів із частотою 1 до загального обсягу тексту), індекс концентрації (відношення кількості слів у словнику з абсолютною частотою 10 і більше до загального обсягу тексту). Саме частотний словник дає можливість об'єктивно визначити ці величини.

Деякі закономірності у функціонуванні частин мови встановлено для твору О. Довженка "Поема про море": "... на іменниковій основі творяться тексти розповідного та описового типу: пейзажі, портретні характеристики. Завдяки іменникам досягається статичність опису, лаконізм, стислість. Прикметники ж увиразнюють ознаки предметів, явищ. На дієслівній основі

---

[37] Статистичні параметри стилів / за ред. *В. С. Перебийніс*.— К.: Наук. думка, 1967.— 260 с.
[38] *Перебийніс В. С.* Статистична стилістика // Українська мова: Енциклопедія / редкол.: *В. М. Русанівський* та інші.— 2-ге вид., випр. і доп.— К.: в-во Українська енциклопедія ім.. М. П. Бажана, 2004.— С. 644.

організовується текст, пов'язаний з описом діяльності людини, різними процесами"[39].

3. ЧС письменника (або конкретного його твору) дає інформацію про інші стилістичні особливості письменника на рівні лексики, наприклад, кількість слів із територіальних чи соціальних діалектів, їхня частота вживання тощо. Скажімо, А. Бєлий уклав частотні списки іменників, прикметників та дієслів на позначення сонця, місяця, неба, повітря, води в поезіях Пушкіна, Баратинського і Тютчева. Після опущення слововживань, характерних усім трьом авторам, оперуючи рештою з них, він показав особливості сприйняття природи кожним з поетів[40].

4. ЧС письменників допомагають встановити (ідентифікувати) авторство творів чи їх фрагментів, оскільки кожен автор має свої так звані "улюблені" слова чи конструкції, які в його творчості мають найвищу частотність. І, навпаки, можна визначити ті слова, які не функціонували в суспільстві у період його діяльності, тому не могли трапитися в його творчості[41].

5. На основі порівняння ЧС різних функціональних стилів, ЧС письменників і т. ін. можна з високою ймовірністю визначити лексичну основу мови, тобто об'єктивно найбільш уживані слова, які слід насамперед засвоїти іноземцеві, що вивчає цю мову. Статистичний підхід до виокремлення словників-мінімумів має довгу традицію в Європі[42]. На жаль, у сучасній лексикографії і дидактичній літературі, послуговуючись назвою "найуживаніша лексика", автори часто не вказують принципів відбору слів до словника, добираючи лексикон інтуїтивно (див., наприклад, праці Сахнів[43]).

---

[39] *Дарчук Н. П.* Статистичні характеристики лексики як відображення структури тексту // Мовознавчі студії.— К.: Наукова думка, 1976.— С. 102.
[40] *Баевский В. С.* Справочнные труды по поэзии Пушкина и его современников.— [Цит. 03 січня 2006].— Доступно з <http://feb-web.ru/feb/pushkin/serial/v91/ v91-065-.html?cmd=p.htm>
[41] От Нестора до Фонвизина. Новые методы определения авторства / Под ред. *Л. В. Милова*.— М.: «Прогресс», 1994.
[42] *Бук С. Н.* Лексична основа української мови: виділення та системно-структурна організація.— Рукопис. Дис... канд. філол. наук: 10.02.01 / Львівський національний університет імені Івана Франка.— Львів, 2004.— С. 14–25.
[43] *Сахно І. П., Сахно М. М.* Словник сполучуваності слів української мови (найуживаніша лексика).— Дніпропетровськ: Видавництво Дніпропетровського ун-ту, 1999.— 544 с.

6. На основі зіставлення частотних словників письменників, що були сучасниками, можна реконструювати особливості мовлення певного періоду, тобто визначити реєстр слів до історичного словника того часу.

Отже, ЧС подає різносторонню статистичну інформацію про текст і має суттєві **переваги** у виявленні особливостей функціонування усіх одиниць лексичного рівня твору у порівнянні з іншими лексикографічними працями, об'єктом опису яких є текст письменника. Тому з метою комплексного квантитативного опису ідіостилю письменника насамперед було вирішено укласти повний ЧС текстів І. Франка (далі — ЧСФ).

### Методика укладання частотного словника текстів Франка

Обсяг спадщини І. Франка за попередніми обчисленнями становить орієнтовно 8 млн. слововживань, то ж у зв'язку зі стрімким розвитком технічних засобів опрацювання мови стає зрозуміло, що вирішити вказане завдання без використання комп'ютера неможливо. Також це зумовлює визначення основних етапів здійснення задуму: опрацювання художньої прози, літературно-критичних праць, наукових розвідок, епістолярію, поезії. У подальшій перспективі можливе й електронне опрацювання рукописів, що, очевидно, є особливо складним для реалізації. ЧС укладається для кожного окремого твору у межах кожного з цих родів літератури.

Зважаючи на складність правописного питання текстів І. Франка, (сам автор у різні періоди творчості писав різними правописами, а сучасні редактори, з метою наближення до сучасного мовлення, вносили ще й свої правки), **джерелами** ЧСФ є академічне Зібрання творів у 50-ти томах[44], а також видання творів, що до нього не ввійшли[45]. Усвідомлюючи неповноту цих джерел, допускаємо ширше охоплення матеріалу, зокрема передбачено опрацювання першодруків та прижиттєвих видань І. Франка (їх розглядаємо у кожному конкретному випадку), зіставлення їх текстів. Так, наприклад, у процесі роботи над ЧС "Перехресних стежок" було зіставлено прижиттєвий першодрук роману 1900 р. та академічне його перевидання 1979 р. У результаті виявлено основні відмінності тексту вказаних видань, зокрема відновлено написання літери "ґ" у власних назвах (*Ваґман, Реґіна, Рессельберґ, Ґенцьо, Ґоттесман*), словах, запозичених з польської, німецької, латинської мов (*ґратулювати, ґуст, ведлуґ, абнеґація, резиґнація, морґ* (міра площі), *ґешефт, ґазета*) і давніших, добре засвоєних у мові, запозиченнях (*ґанок, ґрунт, ґрасувати, ґатунок, ґречно*).

---

[44] *Франко І.* Зібрання творів. У 50-ти томах.— К.: Наукова думка, 1979.
[45] *Франко І.Я.* Мозаїка: Із творів, що не ввійшли до Зібрання творів у 50-ти томах / Упоряд. З.Т.Франко, М.Г.Василенко. – Львів: Каменяр, 2002.— 432 с.

З'ясовано правописні відмінності, які не впливають на статистичну структуру тексту: написання назв національностей з великої літери (*Русини, Жиди, Поляки*); використання "ї" на позначення пом'якшення попереднього приголосного на місці сучасного "і" (*лїкар, усї, неділї, дївчата, молодїж*); використання закінчення Н.в. іменників середнього роду "-нє", "-тє" на місці сучасного "-ння", "-ття" (*напруженє, житє*), використання початкової літери "и" (*иньший*) та інші (*фіртка, ґімназіяльний, історія, репутацию, усьміх*), а також ті, які мають такий вплив. Серед них — написання зворотної дієслівної частки -*ся* з дієсловом окремо (а частки -*сь* — разом), написання часток *б* і *ж* через дефіс з попереднім словом (*коли-б, чого-ж, повинна-б, се-ж*), написання сучасних прислівників, що починаються колишніми прийменниками, окремо (*з далека, до дому, в низу, у двоє, в десятеро, від разу, до схочу, як найшвидше*). Цікаво, що частка -*ся* у повісті трапляється 2496 разів у 1485 дієслівних словоформах — це друге (!) за частотою значення після І / Й.

Для того, щоби результати статистичного опису творів можна було коректно порівнювати між собою, важливо, що би ЧС кожного твору було укладено за єдиною методикою: "відмінності, інколи вельми відчутні, у методиці побудови словників сильно ускладнюють їх коректне порівняння"[46]. Принципи створення ЧСФ розроблено із врахуванням практики укладання трьох ЧС функціональних стилів української мови[47], а також із врахуванням графічних та граматичних відмінностей текстів І. Франка від сучасної літературної мови. Оскільки ці принципи дещо модифікувалися і відшліфовувались у процесі укладання ЧС роману "Перехресні стежки", то приклади, наведені для наочності, взято з цього твору.

---

[46] *Якубайтис Т. А.* О статистических пластах лексики // Вопросы статистической стилистики.— К.: Наукова думка, 1974.— С. 300.

### Принципи укладання частотного словника творів Франка

Як було сказано, ЧС становить собою впорядкований список слів, забезпечений даними про частоту їх вживаності в тексті.

У ЧСФ окремим словом вважаємо послідовність літер (тут апостроф і дефіс розглядаються як літера) між двома пропусками чи розділовими знаками, тому складні числівники виступають як різні слова. Це стосується займенників типу *абихто*, які в непрямих відмінках з прийменником втрачають єдність написання (*аби з ким*). Написання через дефіс розглядаємо як одне слово (*з-поміж, байдужно-спокійний, адвокат-русин* тощо).

ЧСФ подає інформацію про словникові одиниці (тобто леми або слова, зведені до початкової форми) і про словоформи: парадигматичні форми і фонетичні варіанти слів.

Формування ЧСФ здійснено за графічним збігом лем, і кожна частина мови має свою **схему об'єднання словоформ під лемою** (аналогічну, як і в частотних словниках художньої прози, публіцистики, розмовно-побутового стилю).

*Іменник* — до називного відмінка однини зводимо форми всіх відмінків однини та множини. Частоту множинних іменників зводимо до форми називного відмінка множини. Оскільки такі форми іменників як СЕЛЯНИ,

---

[47] *Бук С.* 3 000 найчастотніших слів наукового стилю сучасної української мови / Наук. ред. *Ф. С. Бацевич*.— Львів: ЛНУ імені Івана Франка, 2006.— 192 с.; *Бук С.* 3 000 найчастотніших слів розмовно-побутового стилю сучасної української мови / Наук. ред. *Ф. С. Бацевич*.— Львів: ЛНУ імені Івана Франка, 2006.— 180 с.; *Бук С.* Частотний словник офіційно-ділового стилю: принципи укладання та статистичні характеристики // Лінгвістичні студії: Зб. наук. праць.— 2006.— Випуск 14.— С. 184–188.

РУСИНИ, ЖИДИ, МІЩАНИ і їм подібні позначають осіб і чоловічої, і жіночої статі, їх не об'єднуємо із одниною ані чоловічого, ані жіночого роду.

*Прикметник* — до називного відмінка однини чоловічого роду зводимо відмінкові форми всіх родів в однині та множині, вищий і найвищий ступені порівняння, за винятком суплетивних форм, які зводимо окремо до називного відмінка однини чоловічого роду вищого ступеня, наприклад, БІЛЬШИЙ, НАЙБІЛЬШИЙ зведено до БІЛЬШИЙ. Суплетивні форми прикметників та прислівників не зводились до звичайного ступеня і в більшості вищезгаданих словопокажчиках і словниках.

*Займенник* — відмінкові форми зводимо відповідно до типу відмінювання.

*Числівник* — відмінкові форми зводимо відповідно до типу відмінювання.

*Дієслово* — зводимо до інфінітива синтетичні форми часу (теперішній, минулий і майбутній), форми наказового способу і дієприслівник, а також неособові форми на -но, -то. Аналітичні форми часу вважаємо синтаксичними утвореннями, кожну складову яких зареєстровано як окреме слово.

*Дієприкметник* — до називного відмінка однини чоловічого роду зводимо відмінкові форми всіх родів в однині та множині, оскільки, за І. Вихованцем, розглядаємо його як різновид віддієслівного прикметника із властивими йому основними категоріями (рід, число, відмінок) та типовою синтаксичною роллю означення[48].

*Прислівник* — зводимо вищий і найвищий ступені порівняння, за винятком суплетивних форм.

Лематизацію слів з частками *-бо, -но, -таки, -то* реалізуємо так: самостійні частини мови зводимо до початкових форм (*говоріть-бо* до ГОВОРИТИ, *брешіть-бо* до БРЕХАТИ, *ходи-но* до ХОДИТИ, *колупнули-таки* до КОЛУПНУТИ, …); у службових частинах мови ці частки зберігаємо (ДУЖЕ-ТО, КОЛИСЬ-ТО, АЛЕ-БО…). Як окремі слова залишаємо також випадки

---

[48] *Вихованець І.* Частиномовний статус дієприкметників // Українська мова.— 2003.— № 2.— С. 54–60.

вживання синтетичних особових форм дієслова: ЯКБИ-СТЕ [купували], [потрактував] БИ-С.

В окремих випадках, коли початкову форму слова однозначно відтворити складно, зберігаємо словоформу у тому вигляді, в якому вона функціонує в тексті: ПАНЦЮ (ПАНЦЬО?), пор. також словник З. Великодворської[49].

У ЧСФ намагаємося розмежувати лексичну та лексико-граматичну (іменник МАТИ і дієслово МАТИ) омонімію, зокрема омографи (*наймúти* (дієслово зі значенням "найняти") і *нáймити* (іменник у множині); *гóрод* і *горóд*, *мукá* і *мýка*). У цих випадках для розрізнення в дужках подаємо або вказівку на значення (МІЛЯ (*ім'я*)), або на частиномовну приналежність (МАТИ (*ім.*) і МАТИ (*дієсл.*)). Біля абревіатур та скорочень в дужках вказуємо їх розшифрування: Т. Д. (так далі), О. (отець), Д-Р (доктор) і т.д.

Ті скорочення, які у 50-томнику розшифровано у квадратних дужках (КС[ЬОНДЗ], Т[АК] ЗВ[АНИЙ], ГУЛЬД[ЕНІВ], З[ОЛОТИХ] Р[ИНСЬКИХ]), враховуємо як повну форму (КСЬОНДЗ, ГУЛЬДЕНІВ).

Наголос у словнику подаємо лише в тому випадку, коли він відіграє смислорозрізнювальну роль (СÁМИЙ і САМИ́Й, ВИ́КЛИКАТИ і ВИКЛИКÁТИ див. також попередній абзац) або поданий у тексті ([річ] НАБУТНÁ), оскільки загальна акцентуація мовлення Галичини зламу XIX–XX ст. вимагає окремого дослідження.

До однієї початкової форми (найчастотнішої) зводимо фонетичні варіанти слів, де чергування початкових чи кінцевих літер пов'язане з милозвучністю мови, а саме: дієслова з постфіксами -СЯ / -СЬ; ІТИ / ЙТИ; сполучники ЩОБ / ЩОБИ, І / Й; частки Ж / ЖЕ, Б / БИ; прийменники У / В, З / ІЗ / ЗІ / ЗО, ПІД / ПІДО та деякі інші, а також слова з відповідними префіксами (ВЕСЬ / УВЕСЬ / ВВЕСЬ, ВСЯКИЙ / УСЯКИЙ).

Лексеми ЛЕДВЕ / ЛЕДВО, ТРОХИ / ТРОХА, ТІЛЬКИ / ТІЛЬКО і подібні подаємо в одній словниковій статті, оскільки у прижиттєвому виданні

---

[49] *Великодворская З. Н. и др.* (ред.) Частотный словарь романа Л. Н. Толстого "Война и мир" / М-во просвещения РСФСР. —Тула: Б. и., 1978. — С. 8.

1900 року послідовно вжито форми ТІЛЬКО, СКІЛЬКО та інші. У 50-томнику розрізнення вказаних форм є штучним[50]: авторську форму (ТІЛЬКО, СКІЛЬКО, ЛЕДВО) залишено у прямій мові, а сучасну літературну норму подано в інших випадках.

Натомість форми, які відображають особливості мовлення персонажів, зокрема просторічні, подаємо окремо (АДУКА(Н)Т і АДВОКАТ, ПЕРЕГРАФ і ПАРАГРАФ, КАЗЕТА і ҐАЗЕТА).

Лексеми, написані некириличною графікою лематизуємо відповідно до граматики тієї мови, до якої вони належать. Числа, написані цифрами, вважаємо окремим словом.

Цікаво, що в одному лише романі "Перехресні стежки" зафіксовано 45 слів, записаних цифрами, та 208 слів, написаних латинською графікою: німецькою (87), латинською (55), польською (38), французькою (14), чеською (9), їдиш (4), а також один раз — у контексті: "Та ось поперек його дороги простягається чорна стрічка, закривлена по обох краях обрію, мов велике, плазом покладене S". Серед них також трапляються омоніми: *in* (лат.) й *in* (нім.), *a* (лат.) й *a* (польськ.), німецькі означені артиклі *die* (жіночого роду і множини), латинське *maxima* функціонує один раз як прикметник (жін. роду від *maximus*), а другий раз — як іменник (множина від *maximum*). Цікавим є також міжмовний омонім *на* — прийменник в українській мові та частина єврейського словосполучення *на хайрем* (слово честі).

**Етапи укладання ЧС**

ЧС кожного твору І. Франка укладаємо напіватоматичним способом у декілька етапів:

1. Створення електронної форми тексту шляхом сканування з подальшим детальним його вичитуванням (графічна та граматична специфіка текстів І. Франка вимагає надзвичайно ретельного підходу до цього завдання, оскільки звичайні програми перевірки орфографії не розраховані на тексти західноукраїнського варіанту української мови кінця XIX ст.) та вилученням підсторінкових редакційних приміток. Аналізу підлягають усі слова текстів, включно з написаннями латинською графікою та цифрами.

---

[50] Від редакційної колегії // *Франко І.* Зібрання творів у 50-ти томах.— Т. 1: Поезія.— К.: Наукова думка, 1976. — С. 14–15.

2. Усунення омонімії шляхом додавання до одного із омонімічної пари умовної позначки. Таким чином ці слова стають графічно різними, і програма рахує частоту вживання кожного з них окремо.

3. Автоматичний підрахунок абсолютної частоти кожної словоформи за допомогою спеціально написаної комп'ютерної програми[51]. Результатами цієї операції стають *частотний список словоформ за спадом частот*.

4. Лематизацію, тобто зведення словоформ до словникової форми (початкової форми, леми); наприклад, словоформи АДВОКАТА, АДВОКАТАМ, АДВОКАТАМИ, АДВОКАТИ, АДВОКАТОВІ, АДВОКАТОМ — до леми "АДВОКАТ" здійснюємо напівавтоматично. Повністю автоматичну систему розмітки української мови[52], зорієнтовані на сучасну загальнолітературну норму, у чистому вигляді застосувати до творів І. Франка неможливо через те, що, по-перше, письменник писав західним варіантом літературної мови другої половини XIX – поч. XX ст.; по-друге, навмисне використовував неправильні форми слів у мовленні персонажів; по-третє, його правопис має графічні особливості (наприклад, лїс, усї); по-четверте, немає підсумованих морфологічно неоднозначних словоформ (омонімів) західного варіанту української літературної мови зламу XIX–XX ст., що уможливлює неправильну їх автоматичну лематизацію. Так, скажімо, словоформу ПАНЯ за правилами літературної мови програма мала би вважати іменником IV відміни середнього роду однини в Н. в. (аналогічно до курча, слоня), тоді як

---

[51] Автор-розробник — А. Ровенчак, доцент кафедри теоретичної фізики Львівського національного університету імені Івана Франка.
[52] Корпусна лінгвістика / *В.А.Широков, О.В.Бугаков, Т.О.Грязнухіна* та ін. – К.: Довіра, 2005. – Розділ 5, 6.

І. Франко вживав його як іменник І відміни жіночого роду однини (зі значенням пані, жінка); словоформу *мойого* як слово із прикметниковим закінченням могла звести до *мойий* і т. д.

5. Автоматичний підрахунок абсолютної частоти кожної леми за допомогою спеціально написаної комп'ютерної програми (див. етап 3). Результатом цієї операції стають *ЧС слів за спадом частот*.
6. Розташування всіх зведених лем шляхом сортування в алфавітному порядку. Результатом цієї операції стають допоміжні списки *ЧС слів за алфавітом*.

Таким чином, ЧС кожного окремого твору І. Франка повинен мати три списки: 1) ЧС слів за спадом частот; 2) ЧС словоформ за спадом частот; 3) ЧС слів за алфавітом. Останній виконує допоміжну роль для знаходження слова. Очевидно, що Проект повної квантитативної параметризації текстів Івана Франка триватиме не один рік. До нього залучено студентів філологічного факультету Львівського національного університету імені Івана Франка, зокрема зі спеціальності "прикладна лінгвістика". На сьогодні створено ЧС роману "Перехресні стежки"[53], казок "Вовк війтом", "Мавка", "Лисичка-Кума", "Три міхи хитрощів", "Ворона і Гадюка", які доступні в електронній формі. Повний список слів та словоформ, як частотний, так й алфавітний, а також повні тексти вказаних творів можна знайти на веб-сторінці, присвяченій цьому лінгвостатистичному дослідженню: http://www.ktf.franko.lviv.ua/~andrij/science/Franko/. Визначено основні параметри створення корпусу текстів Івана Франка[54].

На основі ЧС роману "Перехресні стежки" вже зроблено низку лінгвостатистичних досліджень, зокрема проаналізовано частотні розподіли словоформ залежно від кількості складів і фонем, на підставі чого підтверджено закон Менцерата(–Альтманна), за даними розподілу "ранг–частота" розраховано параметри законів Ціпфа і Ціпфа–Мандельброта[55] Наступним етапом роботи над "Перехресними стежками" заплановано аналіз на рівні ієрархії "складне речення ← підрядне речення ← слово", морфологічне теґування тексту, що в перспективі дозволить виконувати згадані вище процедури лематизації автоматично. До речі, під час лематизації словоформ виявлено, що запропонованого О. Демською-Кульчицькою[56] набору теґів морфологічної анотації недостатньо для вичерпного маркування текстів такого типу, як досліджуваний. Зокрема, у вказаній схемі не враховано різних форм займенника *мого / мойого, свого / свойого* і т. ін. Одним зі способів розв'язання цієї проблеми є розширення набору теґів для врахування паралельних словоформ, а також — у перспективі — для позначення різних правописних систем української мови. Такий розширений набір теґів буде корисним як для розмітки сучасних текстів, автори і редактори яких використовують часом дещо різні правописні норми, так і для створення корпусу текстів Івана Франка, укладеного на підставі оригінальних видань.

---

[53] *Бук С., Ровенчак А.* Частотний словник роману І. Франка "Перехресні стежки" // Стежками Франкового тексту (комунікативні, стилістичні та лексикографічні виміри роману "Перехресні стежки") / *Ф. С. Бацевич* (наук. ред), *С. Н. Бук, Л. М. Процак, А. А. Ровенчак, Л. Ю. Сваричевська, І. Л. Ціхоцький*. – Львів: Видавничий центр ЛНУ імені Івана Франка, 2007.— С. 138-369.

[54] *Бук С.* Корпус текстів Івана Франка: спроба визначення основних параметрів // Прикладна лінгвістика та лінгвістичні технології: MegaLing-2006: Зб. наук. пр. / НАН України. Укр. мовн.-інформ. фонд, Таврійськ. нац. ун-т ім. В. І. Вернадського; за ред. *В. А. Широкова*.— К.: Довіра, 2007.— С. 72–82.

**Частотний словник творів І. Франка і корпус текстів І. Франка**

Під корпусом текстів розуміють зібрання електронних текстів, що відповідає вимогам репрезентативності, збалансованості, розміченості (анотованості), стандартності[57].

Корпус текстів І. Франка буде репрезентативним і збалансованим, якщо міститиме усі тексти письменника в електронній формі. Анотація (маркування, теґування, розмітка) — це позначення в тексті морфологічної, синтаксичної, семантичної інформації (теґів). Морфологічне маркування (вказівка на рід, число, відмінок для іменників, час, вид, спосіб ... для дієслів і т. д.) співзвучне з процедурою лематизації у ЧС. Виникає питання, навіщо спочатку лематизовувати словоформи для ЧС (плодити ще один зайвий(?) продукт), якщо пізніше фактично цю ж процедуру доведеться виконувати для корпусу текстів? Чому відразу не протеґувати текст? Відповідь на це питання лежить у площині співвідношення кількості словоформ (обчислених комп'ютерною програмою автоматично) і кількості слововживань (тобто окремих слів у тексті), тобто у площині обсягів необхідної обробки матеріалу. Так, ЧС роману "Перехресні стежки" зафіксував його обсяг 93 885 слововживань, 19 390 різних словоформ і 9964 різних слів. Відповідно можна порахувати, що кількість словоформ майже у п'ять разів менша від кількості слововживань: 93 885 / 19 390 = 4,8.

---

[55] *Buk, S., Rovenchak, A.* Statistical Parameters of the Novel *Perekhresni stežky* (*The Cross-Paths*) by Ivan Franko // Quantitative Linguistics.— V. 62: Exact Methods in the Study of Language and Text.— Berlin; New York, 2006.— P. 39–48.
[56] *Демська-Кульчицька О.* Основи національного корпусу української мови.— К.: Інститут української мови національної академії наук України, 2005.— С. 111–140.
[57] *Meyer C. F.* English Corpus Linguistics: An Introduction.— Cambridge: Cambridge University Press, 2002.— 168 s.; Podstawy językoznawstwa korpusowego / Red. *B. Lewandowska-Tomaszczyk*.— Łódź: Wydawnictwo Uniwersytetu Łódzkiego, 2005.— 306 s; Корпусна лінгвістика / В.А.Широков, О.В.Бугаков, Т.О.Грязнухіна та ін. – К.: Довіра, 2005. – 471 с.; *Демська-Кульчицька О.* Основи національного корпусу української мови.— К.: Інститут української мови національної академії наук України, 2005.— 219 с.

Опрацювавши ці словоформи, вводимо їх разом з початковою формою у словник програми теґування, і вона, натрапивши на цю словоформу в тексті, безпомилково зведе її до правильної леми (про розрізнення омонімів див. вище). Ця процедура також виконує функцію контролю, оскільки дає можливість переконатися, чи результати "ручної" роботи і машини збігаються. Суттєво зазначити, що інтерфейс програми, яка опрацьовує такі складні тексти, як Франківські, обов'язково повинен бути людиноконтрольований, тобто щоби у спірних випадках (наприклад, натрапивши на морфологічно неоднозначні форми) машина ставила питання, а людина, враховуючи контекстний аналіз, сама приймала рішення. Інакше машина помилково присвоїть слову неправильний теґ, а після завершення роботи програми виявити його буде якщо не неможливо, то дуже складно. Слід також зазначити, що "ручний" спосіб розмітки не зовсім вийшов з ужитку, оскільки хоча й вимагає непорівнянно більше часу, проте вважається більш якісним. Наприклад, Національний корпус української мови Інституту української мови НАНУ на даному етапі також тегують вручну[58].

Отже, квантитативна параметризація текстів Івана Франка, що поетапно реалізовується у створенні ЧС творів письменника, дає якісно новий матеріал для лінгвостатистичного дослідження його стилю. Така праця виявить цінні дані для укладання словника української мови зламу XIX–XX ст. на зразок Словника української мови XVI — першої пол. XVII ст.[59], оскільки письменник послуговувався багатьма функціональними стилями у різних царинах людського духу, зокрема художнім, публіцистичним, науковим, епістолярним.

Реалізація цього проекту, окрім самостійної теоретичної й практичної ваги, може слугувати одним з етапів роботи над словником мови письменника (етап визначення реєстру слів), а також над створенням повного корпусу текстів І. Франка.

## Quantitative parametrization of texts written by Ivan Franko: An attempt of the project

*Solomiya Buk*

In the article, the project of quantitative parametrization of all texts by Ivan Franko is manifested. It can be made only by using modern computer techniques after the frequency dictionaries for all Franko's works are compiled. The paper describes the application spheres, methodology, stages, principles and peculiarities in the compilation of the frequency dictionary of the second half of the 19th century – the beginning of the 20th century.

The relation between the Ivan Franko frequency dictionary, explanatory dictionary of writer's language and text corpus is discussed.

---

[58] *Демська-Кульчицька О.* Основи національного корпусу української мови.— К.: Інститут української мови Національної академії наук України, 2005.— С. 111–140.

[59] Словник української мови XVI — першої пол. XVII ст. / НАН України, Ін-т українознавства ім. І. Крип'якевича.— Вип. 3.— Львів, 1996.— 251 с.